# Protein-Ligand Complex Generator & Drug Screening via Tiered Tensor Transform

**Jonathan P. Mailoa,[1][†][*] Zhaofeng Ye,[1][†] Jiezhong Qiu,[1] Chang-Yu Hsieh,[1] and Shengyu Zhang[2][*]**

1) Tencent Quantum Laboratory, Tencent, Shenzhen, Guangdong, China
2) Tencent Quantum Laboratory, Tencent, Hong Kong SAR, China

[†] These authors contributed equally to this work

[*] corresponding author: jpmailoa@alum.mit.edu, shengyzhang@tencent.com

# Protein-Ligand Complex Generator & Drug Screening via Tiered Tensor Transform

**Abstract**

The generation of small molecule candidate (ligand) binding poses in its target protein pocket is important for computer-aided drug discovery. Typical rigid-body docking methods ignore the pocket flexibility of protein, while the more accurate pose generation using molecular dynamics is hindered by slow protein dynamics. We develop a tiered tensor transform (3T) algorithm to rapidly generate diverse protein-ligand complex conformations for both pose and affinity estimation in drug screening, requiring neither machine learning training nor lengthy dynamics computation, while maintaining both coarse-grain-like coordinated protein dynamics and atomistic-level details of the complex pocket. The 3T conformation structures we generate achieve significantly higher accuracy in active ligand classification than traditional ensemble docking using hundreds of experimental protein conformations. Furthermore, we demonstrate that 3T can be used to explore distant protein-ligand binding poses within the protein pocket. 3T structure transformation is decoupled from the system physics, making future usage in other computational scientific domains possible.

# Introduction

Structure generation and optimization is an essential topic in the field of life and physical sciences. The applications of generative model algorithm in these fields are diverse, ranging from precipitation nowcasting[1] and airfoil aerodynamics optimization[2–4] down to the optimization of optical nanostructures[5–7], electrode microstructures,[8] microfluidic devices,[9,10] and material design.[11–13] Structure generation capability is even more prevalent and important for drug discovery applications,[14–20] where protein and drug candidate small molecule (ligand) structures are of interest in the determination of suitable target-specific drug molecules. Recent development in the deep learning community has enabled generative model algorithms to more accurately predict single protein structures,[21,22] with a notable recent example being the success of AlphaFold 2 in the CASP14 competition.[23] Other notable examples are similarly focused on either the generation of small molecules with desired properties[24,25] or on the high-efficiency sampling for protein structures (RiD, RIP, L-RIP, TAMD, etc.).[26–31]

Targeting a specific pocket of a protein with a ligand poses an additional challenge because the compound is composed of multiple interacting structures. Finding the binding pose of a flexible ligand molecule when it is docked onto a flexible protein receptor pocket is a daunting task due to the degree of flexibilities built onto this two-system complex. The most common method (with lower computational cost) is to freeze the target protein receptor pocket and use a docking software such as AutoDock Vina,[32] Smina,[33] and Glide[34] to attach the ligand molecule onto the rigid protein pocket[35] and generate multiple candidate ligand docking poses. Unfortunately it is known that ligands docked onto a rigid protein pocket are not representative of how ligands look like in a real protein-ligand complex and are difficult to use for active ligand classification,[36] partly because the protein structure is flexible and can undergo intrinsic or induced conformational changes.[37,38] The state-of-the-art solution to this problem is to perform ensemble docking, where multiple structures of the same protein pocket (usually $10^2 - 10^3$ structures) are either obtained experimentally or generated through long molecular dynamics simulations (MD with millisecond-long simulations, which correspond to ~$10^{12}$ MD time steps)[36,39] and other generative methods;[40] the ligands of interest are then docked to all these protein structures to generate multiple conformations of protein-ligand complexes. It is worth noting that the protein structure generation is done with no regard to the specific ligand existence in the protein pocket, as it remains difficult to simultaneously modify the protein and ligand structures due to the combined degree of flexibilities in the protein-ligand complex. To the best of the authors' knowledge, MD simulation of the entire protein-ligand complex structure remains the only reliable way to sample and generate diverse complex conformations of entire protein-ligand pockets.[41,42] Unfortunately, it is computationally expensive to do so due to the slow dynamics of the protein, and this kind of process is usually reserved for the last step of a drug screening workflow such as the free energy perturbation (FEP) approach, where 2-4 ligands can be scored per day on a 4-GPU server,[43,44] although there is an attempt to do this over larger scale for the initial screening steps.[45] We note that there are also meta-dynamic physical pathway methods to bias the MD simulation of protein-ligand pockets aiming to accelerate non-equilibrium MD structure sampling such as funnel-metadynamics (FM), umbrella sampling (US), and steered molecular dynamics (SMD).[46–50]

In this work we propose the tiered tensor transform (3T) algorithm, a general framework to generate diverse physical or biological conformation structures of a multi-scale complex system for optimization purposes. We demonstrate the usage of this algorithm on the problem of protein-ligand pocket complex conformation generation by simultaneously generating multiple conformations of the entire protein-ligand complex. 3T requires one example of the structure to optimize as its initial starting point (**Figure 1a**), which is then segmented into local groups in a hierarchical manner into several smaller micro-groups and larger macro-groups as appropriate. In the context of physics or life sciences, connectivity-based segmentation is the most appropriate choice. We segment protein-ligand complexes based on rotatable bonds, protein residues, and protein secondary structures (**Supplementary Figure 1**). We then apply multiple structure tensor transformations on these micro-groups and macro-groups in a hierarchical manner, enabling both macro-scale and micro-scale optimizations which more effectively escape local energy minima and sample multiple and diverse protein-ligand pocket complex conformations. These hierarchical tensor transformation parameters are updated based on the cost function gradient (force field energy, see **Figure 1b**). While our approach is neither a machine learning (ML) approach nor an MD approach, it extensively uses tools commonly found in deep learning (PyTorch[51]) in its structure transformation part and MD (atomistic force field) in its structure evaluation.[52–55] This modularity makes it possible to apply 3T-like algorithm for other types of complex multi-scale structure generation and optimization purposes if a differentiable domain-specific structure evaluation cost function is available (especially helpful in physics-informed deep learning work).[56] 3T method enables full flexibility for all of the protein backbones and sidechain groups during pocket structure generation (**Figure 1c**), unlike semi-flexible protein-ligand docking methods available in state-of-the-art tools such as smina, rDock, GOLD, FLAP, GRID, and ICM which only allow for select protein sidechain dihedral rotation.[33,57–61] These semi-flexible tools' complexity varies, ranging from only allowing rotations of –OH and $-NH_3$ groups of the sidechain[60] to going through most available sidechain rotations exhaustively.[33] The full protein flexibility of 3T method, while theoretically should be much more computationally expensive than the semi-flexible methods, is possible because 3T quickly eliminates energetically unfavourable protein-ligand conformations through a coarse-grain-MD-like hierarchical structure transformations. We utilize 3T to generate protein-ligand co-crystal structure conformations for three representative protein targets, i.e. cyclin-dependent kinase 2 (CDK2), heat stock protein 90 (HSP90) and coagulation factor X (FXa). This 3T structure generation can be performed with more than 80× lower computation cost *vs* comparable MD simulation. We further demonstrate that ten of these 3T protein-ligand complex conformations are superior when used to identify active and decoy ligands of a protein target in DUD-E and DEKOIS 2.0 datasets when compared to state-of-the-art ensemble docking procedure performed on hundreds of experimentally obtained rigid protein host conformations,[62,63] in part because we also have access to the 3T pocket energy landscape near the binding pose local energy minimum structures. Finally, we demonstrate 3T's ability to explore both large protein and ligand conformational space, including exploration of alternative binding modes.

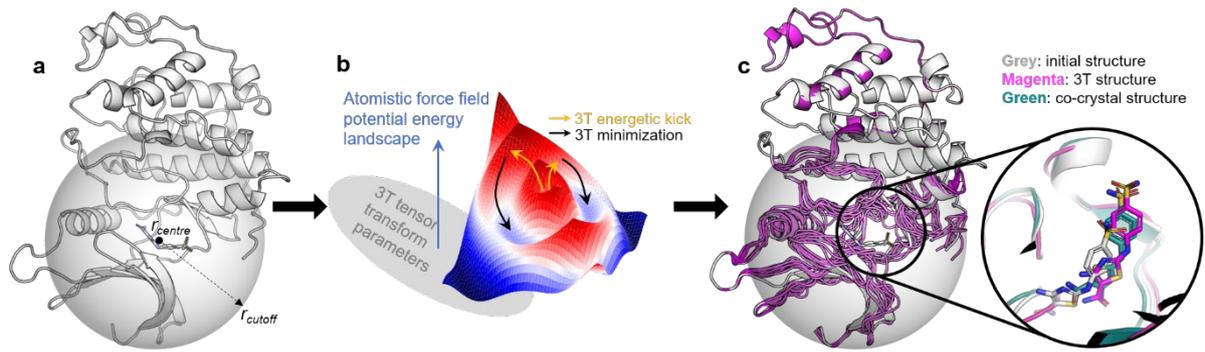

**Figure 1 | Flowchart of 3T structure generation and analyses. a)** Initial protein-ligand pocket volume definition. **b)** 3T energetic kick and optimization in the tiered tensor transform parameter hyperspace. **c)** Protein-ligand pocket structures generated by 3T (only one ligand pose shown for clarity), compared to the initial docked structure and the corresponding experimental protein-ligand co-crystal structure.

## Results & Discussion

### Tiered Tensor Transform Algorithm

In principle, 3T is a generative algorithm which works by transforming an existing structure through multiple scales of tensor transformations, restrained by physics-based cost function to keep the generated multi-scale transformed structures physically realistic, and functionally relevant or optimized. There are three major components in 3T: hierarchical structure segmentation, hierarchical tensor transformation assignments, and differentiable structure evaluation cost function.

Hierarchical structure segmentation is necessary because we would like to enable local transformation operations to be performed on our initial structure. It is known that protein-ligand complexes have a large number of high-dimensional local energy traps, making it more practical to perform ligand docking on rigid protein structures[35] or at most semi-flexible protein structures.[59,60] Performing structure optimization directly on the atomic coordinates of the protein and ligand atoms will only produce relatively small atomic movements. Structure segmentation eliminates some of this problem by grouping locally connected atoms into separate groups which move in a coordinated manner. This grouping is like what is done in coarse-grained molecular dynamics,[64] but 3T maintains its full atomistic level detail. The grouping is also done hierarchically in order to enable different levels of coordinated movements, including the protein backbone. For our protein-ligand pocket complex generation, we segment the atoms onto 2 segmentation levels: micro-groups and macro-groups. Atoms in a pocket (within a cutoff radius $r_{cutoff} = 20\text{Å}$ from the protein-ligand complex centre $r_{centre}$) are considered movable, which are then segmented into separate micro-groups: (1) for protein atoms, the micro-groups are segmented based on their amino acid residue and on backbone/sidechain distinction; (2) for ligand atoms, the micro-groups are segmented based on their rotatable bonds. These micro-groups are then grouped further into separate macro-groups: (1) protein micro-groups in each flexible loop secondary structure (**Methods**) receive separate macro-group assignment while the remaining micro-groups (helixes and sheets) are not assigned into any macro-group; (2) all ligand micro-groups are assigned into one ligand macro-group. This completes our two-level 3T hierarchical segmentation (**Supplementary Figure 1**). In principle, more hierarchical segmentation levels can be used depending on the physical nature of the system under consideration. For example, if a ligand segment has no rotatable bond but more conformations are desired (cyclohexane groups being an example), it will be a good idea to add different types of micro-group segmentation hierarchy to enable fast 3T multi-scale conformation sampling for the sub-structures of interest.

Afterward, we assign separate hierarchical local structural tensor transformations on each micro and macro-group (**Figure 2**, and a more visual version in **Supplementary Figure 2**). Atoms in the same micro-group *i* are transformed using rotation or translation tensor transformation parameters $\theta_{R,i}$ and $\theta_{T,i}$ respectively. $\theta_{R,i}$ and $\theta_{T,i}$ each has 3 scalar parameters corresponding to rotation angle around and translation along the *x*, *y*, and *z* axis originated on the micro-group centre. These tensor transformations give movable atoms additional coordinated rotation and translation degrees of freedom in addition to the individual atom translations during 3T optimization. We also apply a special axis rotation transformation $\theta_{A,i}$ (represented by 1 scalar parameter) on micro-groups which only has one rotatable bond (anchor), allowing these micro-groups to freely rotate around the anchor. Larger-scale 3T coordinated movement is also enabled by employing coordinated rotation and translation on each macro-group *j*, using the transformation parameters $\theta_{R,j}$ and $\theta_{T,j}$ respectively (containing 3 scalar parameters each like their micro-group counterparts).

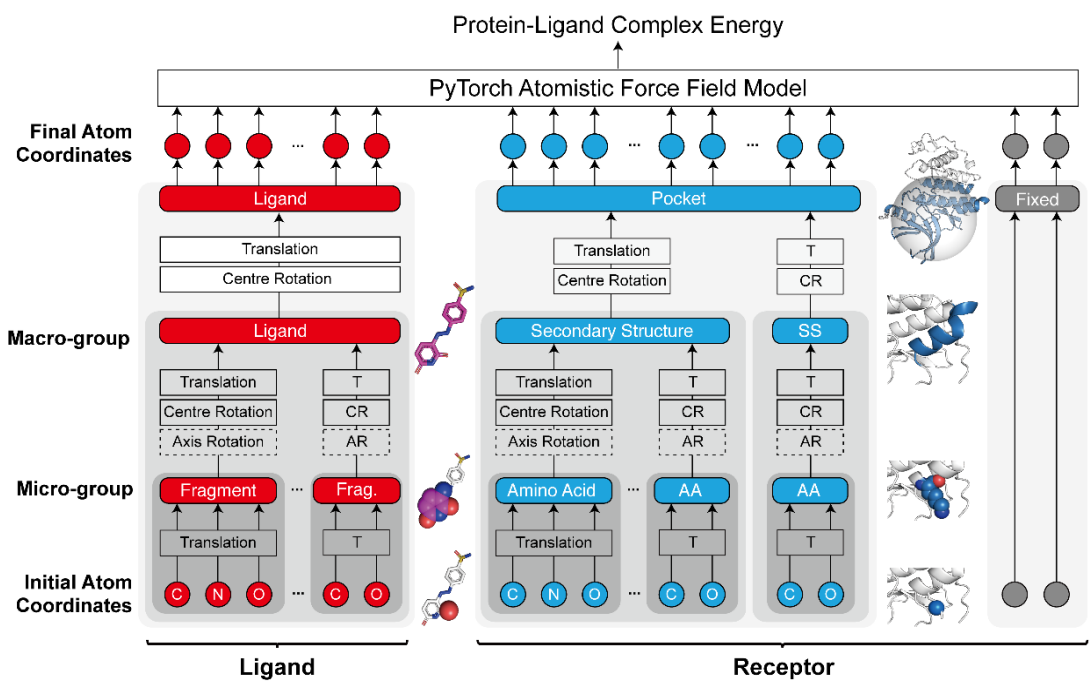

**Figure 2 | Schematic of the 3T scheme.** Protein-ligand complex atoms are separated into fixed (grey circles) and movable atoms (red/blue circles). The movable atoms undergo several layers of multi-scale hierarchical tensor transformations (individual atom translation, sidechain micro-group rotation around rotatable bond axis, micro-group Cartesian axis rotation and translation, and macro-group Cartesian axis rotation and translation). Micro-group axis rotation (dashed boxes) around the rotatable bond is only available for some micro-groups such as protein sidechains and ligand edge fragments. Instead of directly optimizing the final atomic coordinates using atomistic force field, we optimize the 3T tensor parameters and the initial movable atom coordinates during the PyTorch cost function gradient backpropagation. Computationally this will look like a typical deep learning training procedure. See **Supplementary Figure 2** for a better visual of the hierarchical structure transformations.

After these hierarchical segmentation and transformation assignments, the 3T optimization procedure itself is relatively straightforward using a differentiable structure evaluation cost function. We start with the initial fixed and movable atom coordinates $\vec{r}_f$ and $\vec{r}_{m,init}$, pass $\vec{r}_{m,init}$ through the multiple stages of tensor transformations governed by the transformation parameter $\theta$'s above, and calculate the final movable atom coordinates $\vec{r}_{m,final}$. We implement both tensor transformation and atomistic force field model in PyTorch (**Methods**) to calculate the total system energy based on $\vec{r}_f$ and $\vec{r}_{m,final}$. PyTorch is used instead of LAMMPS[65] because PyTorch autograd can perform flexible automatic differentiation to accumulate gradients on 3T multi-scale transformation parameters; this autograd capability is not available in LAMMPS. Using this force field energy as our primary cost function, backward propagation updates the 3T model's adjustable parameters $\vec{r}_{m,init}$ and $\theta$'s. Simply put, we have converted the structure generation cost function calculation from the original (**Equation 1**) to a 3T version (**Equation 2**):

$$C_{E,original} = E_{FF}(\vec{r}_f, \vec{r}_{m,init}) \quad (1)$$

$$C_{E,3T} = E_{FF}\left(3T(\vec{r}_f, \vec{r}_{m,init}, \theta_{R,i}, \theta_{T,i}, \theta_{A,i}, \theta_{R,j}, \theta_{T,j})\right) \quad (2)$$

where $C_E$ is the cost function of the protein-ligand complex, $E_{FF}$ is the atomistic force field energy function, and $3T$ is the hierarchical tensor transformation function illustrated in **Figure 2**.

During the optimization, we first start by optimizing the ligand within rigid protein pocket for $n_{step}$ = 200 steps. We then apply an energetic kick in the system to start our protein-ligand complex pocket conformation generation by initializing small random values on the micro-group $\theta$'s. This energetic kick distorts the structure to a high-energy state, which then gets minimized by 3T over $n_{step}$ = 2000 backward propagation steps. Different random number seed will generate different 3T energetic kick and final structures (**Methods**). Computation-wise, this minimization process looks like a standard deep learning training (with no classification label or regression target in its cost function). While the number of optimizable parameters increases when 3T structural transformation is applied, it is significantly easier to escape local energy minimum and faster to reach lower-energy equilibrium because we have additional coordinated micro- and macro- degrees of freedom.

**Evaluation of Generated Protein-Ligand Pocket Structures**

For the conformation structure generation quality assessment, we compare these 3T structures with available experimental co-crystal structures for specific protein-ligand complexes. In this work, we generate 3T protein-ligand complex conformations for three different proteins: CDK2, HSP90, and FXa. CDK2 pocket is a flexible deep hydrophobic cavity, HSP90 pocket is surrounded with two long alpha helices which are known to take several major conformation changes upon different ligand binding,[66] while FXa active site is a flexible shallow hydrophobic groove. For each of these proteins, we extract one protein structure example from the Protein Data Bank (PDB): 1fin, 1uyg, and 1ezq respectively (see **Figure 3a**).

We first perform individual cross-docking structural analysis on 3T-generated CDK2 conformations as an example. The ligand from 4ez3 PDB co-crystal structure is cross-docked onto our 1fin protein using smina[33] to obtain the initial structure. We calculate the root mean squared displacement (RMSD,[40] see **Methods**) of this cross-docked ligand when compared to the original 1fin co-crystal ligand, producing $RMSD_{init}$ = 3.89 Å. The larger this value is, the farther the predicted cross-docked structure is from the real co-crystal. We then transform this initial structure through 3T (see **Methods**), producing new protein-ligand complex conformation. It can be seen from **Figure 3b** that the 3T ligand conformation matches the original 1fin co-crystal ligand conformation better than the initial structure, with smaller $RMSD_{3T}$ = 1.86 Å and protein binding sites which are more correctly attached to the ligand compared to the initial smina structure with the rigid 1fin protein. We correspondingly have RMSD improvement $\Delta RMSD = RMSD_{3T} - RMSD_{init}$ = -2.03 Å . This improvement is induced by protein backbone conformation change (red circle), which becomes closer to the actual 4ezq co-crystal protein and pushes on the ligand slightly, and can be seen in more detail in **Supplementary Figure 3a**.

We further analyse our 3T CDK2 protein conformations compared to what will be found across known co-crystal structures as well as MD simulations. 373 co-crystal CDK2 protein structures are extracted from the PDB website. 500 protein structures are extracted every 1 ns interval from an MD simulation of the 1fin protein-ligand structure in water (see **Methods**). Finally, 90 protein structures are extracted from 3T-generated structures originating from 1fin smina re-docked initial structures (new conformations can be generated by simply changing the random number seed of 3T energetic kick). Principal component analysis (PCA) of the protein backbone is commonly used to assess protein conformation diversity,[67] and is shown in **Figure 3c** (see **Methods**), showing that the co-crystal structures span a much more diverse range of protein backbone conformations due to the diverse set of co-crystal ligand chemistry geometries found in nature. The principal components (PC) of ligand-specific conformations from the MD occupies a fraction of the co-crystal PC sub-space, and crucially the ligand-specific PC from 3T-generated conformations occupy mostly the MD conformations' PC sub-space outside of the irrelevant general co-crystal PC sub-space. This shows that 3T protein conformations are ligand-induced and more specific compared to general protein-ligand co-crystal structures. 3T final structures rely on energy minimization scheme, so it does not occupy the entire PC space of the 1fin MD conformations which is done at the room-temperature. Similar individual ligand pose and protein microstate examples for HSP90 and FXa pocket conformations are available in **Supplementary Figure 3b-e**.

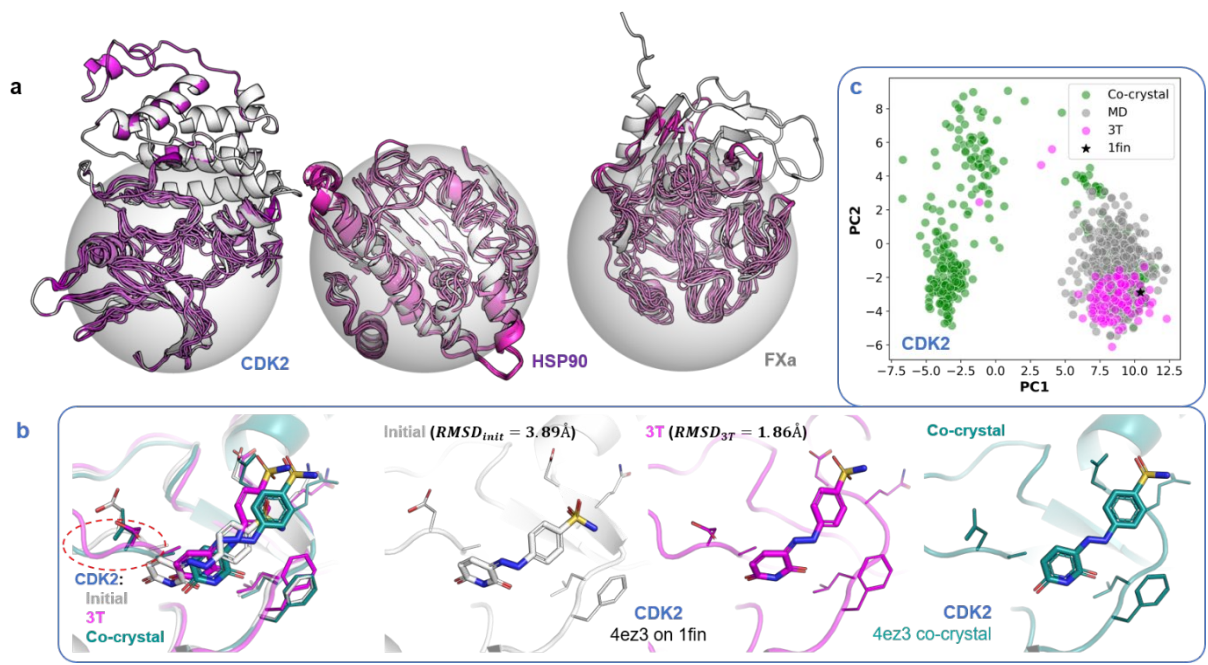

**Figure 3 | Protein-ligand complex conformation generation using 3T. a)** Example of ligand-dependent pocket conformations for three protein structures generated in this work, with CDK2 being the most flexible and HSP90 being the most rigid among the three. Ligand geometries are hidden for clarity. **b)** Overlaid visual comparison of cross-docking on CDK2 protein structure (ligand structure from 4ez3 PDB cross-docked on protein structure from 1fin PDB) using 3T and standard rigid protein docking, compared to the ground truth 4ez3 co-crystal structure. The red circle indicates the protein backbone structure which is correctly transformed by 3T and has become significantly more similar to the co-crystal, which is then responsible for pushing the cross-docked 3T ligand closer to experimental co-crystal ligand pose compared to the initial cross-docked structure. Individual structures (grey: initial, magenta: 3T, green: co-crystal) are available for clarity purposes, showing that 3T structure has a smaller RMSD and a better match (ligand pose and protein binding sites) with the experimental co-crystal. **c)** CDK2 protein backbone micro-state comparison using PCA for 3T-generated structures (re-docking of 1fin PDB) compared to 1fin pocket structures generated using 500 ns protein-ligand complex MD and to all known experimental CDK2 co-crystal structures. The MD only occupies a fraction of the entire CDK2 PC sub-space because there are several major ligand-dependent protein conformations available for CDK2. 3T correctly occupies only the sub-space corresponding to that of 1fin MD and does not occupy the remaining subspace which are not physically accessible by 1fin protein-ligand complex. The experimental 1fin co-crystal structure is shown as the black star. If a semi-flexible cross-docking ensemble is performed using one initial co-crystal structure, the PC will only show up as a single dot here because the protein backbone cannot move, unlike the 3T and MD methods.

## Applicability in Active Ligand Classification

We further evaluate the practical utility of our 3T complex conformations for active ligand classification, compared to complex pocket structures obtained using conventional methods. It has recently been shown that docking potential drug candidate ligand molecules onto a single rigid protein pocket is insufficient for the purpose of active ligand classification.[36,39] In fact, the ligands need to be docked onto hundreds of distinct rigid conformations of the target protein pocket.[36,39] Simple docking score evaluation is insufficient and an ML model needs to be built on top of the ensemble docking scores to obtain a decent active ligand classifier.[39] These varieties of protein conformation structures are difficult or expensive to obtain, and hence ligand (*A*) is often docked onto non-matching rigid protein structure (*B*) taken from a different experimental protein-ligand (*B-C*) complex, or onto rigid protein structure (*D*) generated from lengthy MD of the protein pocket in a solvent. On the other hand, the 3T structure generation enables us to generate ligand-dependent protein-ligand complex pocket conformations explicitly tailored to each protein-ligand pair.

We demonstrate this versatility by performing ML-assisted "ensemble docking" similar to that performed by Ricci-Lopez *et al*.[39] In the prior work, ligand docking was performed onto different number of rigid protein structures depending on the dataset (CDK2: 402, HSP90: 64, FXa: 136). For each ligand in each protein host dataset we perform a molecular cross-docking onto their respective protein target structures using smina[33] to obtain one initial structure for 3T generation (**Methods**). For each initial protein-ligand complex, we generate 10 conformations using 3T energy minimization. In this work, we simply generate ten 3T conformation structures for each ligand docked onto one rigid protein of CDK2, HSP90, and FXa. We adopt the identical procedure of 30×4-fold cross validation (30×4cv) and gradient boosting trees (GBT) classifier algorithm which was used in prior work to ensure that we only compare the conformation feature quality and not the classification method being used.[39] We also note that 3T generates not only the protein-ligand complex pocket conformations, but also the potential energy landscape surrounding the local energy minimum during its structure optimization procedure which can be used as additional features. We hypothesize that it is not simply the shape of the protein-ligand pocket structure (e.g. docking score) which determines how likely it is for a ligand to bind onto a target protein pocket, but also how accessible such protein-ligand pocket energy minimums are (energy barrier landscape surrounding the local energy minimum), as can be seen in **Supplementary Figure 4**. This feature extraction procedure and the subsequent 30×4cv classification process are shown in **Figure 4a**, where for each of the ten 3T conformations we generate for each protein-ligand complex, we extract not only the docking scores but also the protein-ligand binding formation energy $\Delta E = E_{complex,3T} - E_{ligand,3T} - E_{protein,init}$ throughout the 3T optimization process (**Methods**). Due to the large protein size, CDK2 protein-ligand pocket conformations are re-generated with $r_{cutoff} = 25\text{Å}$ for this classification work while $r_{cutoff} = 20\text{Å}$ is kept for both HSP90 and FXa (**Supplementary Table 1**).

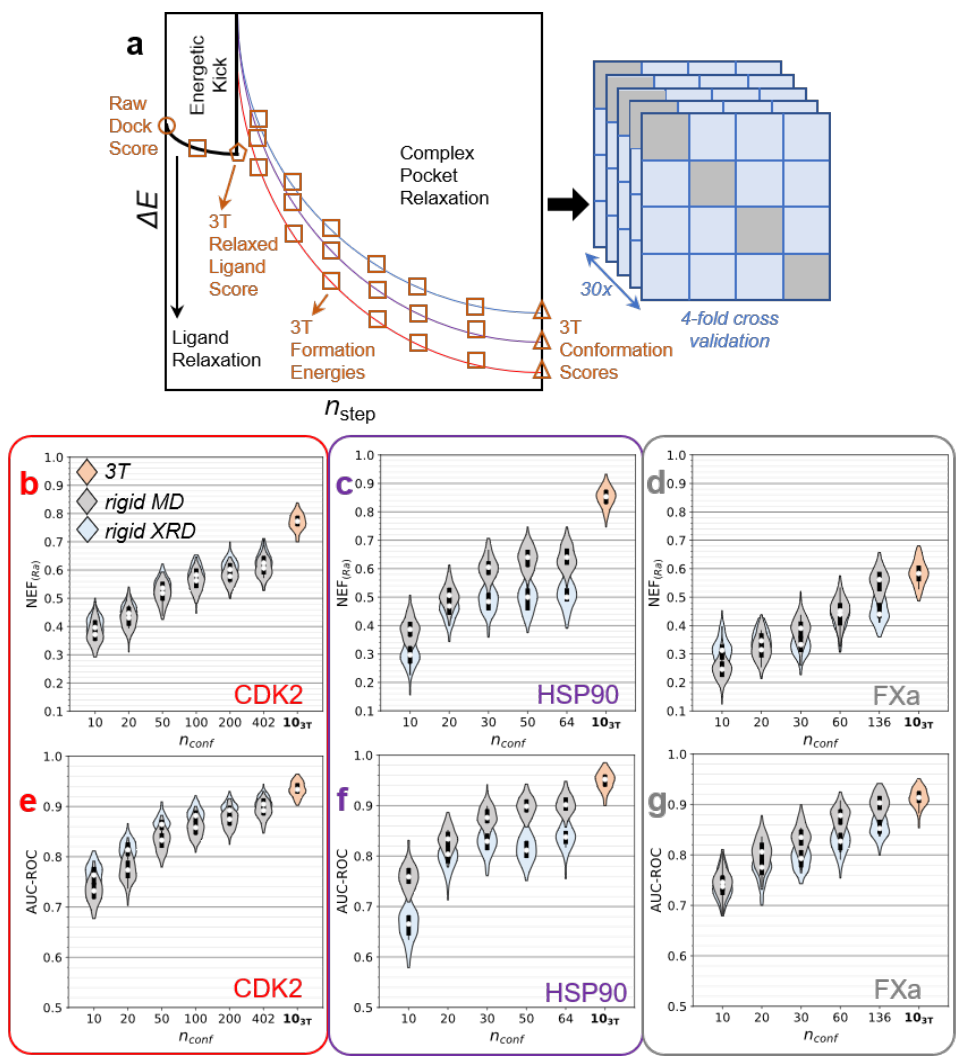

**Figure 4 | Active ligand classification using 3T-generated conformations. a)** 3T conformation feature extraction process (pocket cross-docking scores and formation energies $\Delta E$) and subsequent 30× 4-fold cross validation (with GBT classifier). **b-d)** The $NEF_{R_a}$ classification metric is shown for different number of CDK2, HSP90 and FXa pocket conformations respectively. Similarly, the AUC-ROC metric is shown in **e-g)**. The three proteins differ in structure flexibility, with CDK2 and FXa being dominated by flexible loops and HSP90 being dominated by alpha helixes and beta sheets. We see that the features from ten 3T ligand-dependent pocket conformations generated from one experimental protein conformation are equivalent or better than features from significantly larger number of rigid experimental X-ray diffraction protein conformations (rigid XRD) or simulated MD conformations (rigid MD).

| Protein | Active Ligands | Metric | 3T Classifier | Rigid MD Classifier | Rigid XRD Classifier | $n_{conf,3T}$ vs $n_{conf,MD-XRD}$ |
|---|---|---|---|---|---|---|
| CDK2 | 442/3764 ($R_a = 0.117$) | $NEF_{R_a}$ | **0.771 ± 0.030** | 0.608 ± 0.033 | 0.624 ± 0.039 | (10 + 2) vs 402 |
| | | AUC-ROC | **0.935 ± 0.014** | 0.892 ± 0.017 | 0.904 ± 0.015 | |
| HSP90 | 298/2452 ($R_a = 0.122$) | $NEF_{R_a}$ | **0.851 ± 0.035** | 0.640 ± 0.042 | 0.505 ± 0.046 | (10 + 2) vs 64 |
| | | AUC-ROC | **0.949 ± 0.018** | 0.903 ± 0.019 | 0.836 ± 0.024 | |
| FXa | 298/7191 ($R_a = 0.040$) | $NEF_{R_a}$ | **0.584 ± 0.043** | 0.554 ± 0.046 | 0.452 ± 0.044 | (10 + 2) vs 136 |
| | | AUC-ROC | **0.913 ± 0.018** | 0.902 ± 0.021 | 0.855 ± 0.021 | |

**Table 1 | 3T active ligand classification metrics using the GBT classifier across 3 different protein hosts.** Generally, the more $n_{conf}$ used for a standard ensemble cross-docking classifier, the better the classifier will be. We show that 3T conformation classifiers (10 + 2 = 1 initial + 1 with relaxed ligand + 10 energetic kick conformations) consistently outperform rigid protein conformation classifiers across the 3 different protein hosts even though the standard conformation classifiers use significantly more rigid experimental protein conformation structures.

To enable fair comparison with existing work and direct comparison between different dataset sample distributions, we use the metric area under the curve – receiver operating characteristics (AUC-ROC) and normalized enrichment factor $NEF_\chi = a_s/min(\chi m, a)$ where $m$ is the total number of ligands in the dataset, $\chi$ is the top fraction of the ranked ligands to be selected (set to $\chi = R_a = a/m$), $a$ is the total number of true active ligands, and $a_s$ is the total number of the chosen $\chi m$ ligands which are true active ligands.[39,68] As we would like to investigate how useful our protein-ligand complex pocket conformations are compared to experimentally obtained protein conformations, we re-calculate $NEF_{R_a}$ of the prior work[39] for the three proteins for different number of rigid experimental X-ray diffraction protein conformation hosts (rigid XRD).[39] In addition, we also perform long MD of the proteins in water, extract the structures as shared rigid protein hosts for all of the ligand dockings, and calculate the corresponding classification metric for this MD-based reference (rigid MD) similar to another prior work.[36] We note that these MD structures are not ligand-dependent because performing individual MD for each explicit protein-ligand pair (holo-MD) in this work will be computationally prohibitive. The AUC-ROC result supporting our hypothesis is shown in **Figure 4b**, where we show that an 'ensemble-docking' CDK2 active ligand classifier built using ten 3T protein-ligand complex conformations significantly outperforms an ensemble-docking CDK2 active ligand classifiers built using either 402 rigid protein conformation hosts (both rigid MD and rigid XRD). We similarly outperform 64 HSP90 and 136 FXa rigid protein conformation classifiers using our respective ten 3T conformations (**Figure 4c-d**, **Table 1**). These classifiers' $NEF_{R_a}$ metrics are consistent with AUC-ROC metrics (**Figure 4e-g**) showing 3T conformations significantly outperforming their rigid conformation counterparts, further demonstrating the classification utility of our 3T conformations. The performance improvement is especially big for HSP90, which is the least flexible protein pocket structure among the three. We also show that constraining or eliminating protein structures' flexibility during the 3T conformation generation will significantly degrade 3T classifier performance (**Supplementary Figure 5**, **Supplementary Table 1**).

While it may seem non-intuitive that a classifier built with the number of conformation structures $n_{conf,3T}$ = 10 which originates from $n_{conf}$ = 1 protein host conformation structure achieves similar or better classification results than a classifier built using a large number of experimental protein host conformation structures ($n_{conf}$ = 64–402), we note that the 3T structures are ligand-dependent and their features contain more information (**Methods**). We further note that while 3T conformations are unique and random for each ligand (making classifier feature usage more difficult to justify), we mitigate this problem by ensuring that the same 3T random seed is used across different ligands. This means the ligands share almost identical initial protein structural distortion during the 3T energetic kick process, before the structures get relaxed into their final protein-ligand complex conformation geometries. Finally, we note that 3T classifiers with $n_{conf,3T}$ = 4 is in fact enough to outperform both rigid MD and rigid XRD-based classifiers (**Supplementary Figure 4**).

In addition to being significantly more accurate and taking significantly less experimental resources than conventional approaches, 3T also takes significantly less computation resources than holo-MD or exhaustive smina semi-flexible docking approach[33] (see **Supplementary Table 2**), although this prototype version of 3T is still slower than the lightweight semi-flexible docking approach such as rDock which only allows limited –OH and –NH$_3$ group rotation on the sidechains.[60] If the protein-ligand complexes are generated using holo-MD, 3T structure generation is computationally cheaper than holo-MD by more than 80× (aggressive MD assumption, see **Methods**). This holo-MD approach for each ligand is computationally intractable, and one way to reduce this cost is by performing a protein pocket MD and sharing the compute cost across all ligands of interest prior to rigid-protein docking (rigid MD). Under this rigid MD docking scenario, it is difficult to achieve better active ligand classification performance compared to the rigid protein docking using experimental structures especially when there are multiple major protein conformations such as CDK2 (**Figure 3c**). See **Supplementary Flowchart 1** for more detail on the differences between these methods.

## Sampling of Distant Poses and Pockets

We also demonstrate 3T's capability to generate significantly more difficult conformations which are very far away from the initial structure. This capability is useful for exploring intermediate metastable ligand conformations in different and relatively far protein pockets (~10Å away). These tasks require significantly more aggressive 3T energetic kick and optimization workflow (see **Supplementary Information** for details) than the gentle settings we have used in the previous sections of this manuscript for large scale virtual screening. In **Figure 5**, we utilize 3T to investigate the conformational changes of both the ligand (alprenolol) and GPCR protein (beta-2 adrenergic receptor, ADRB2). The ligand binds to different protein pockets located at the extracellular vestibule, which leads to the entrance of the orthosteric binding site, as previously reported (**Figure 5a**).[69] We show how these more aggressive 3T settings can enable an alprenolol molecule originally sitting in its orthosteric binding site on ADRB2 to jump and explore temporary binding modes at the extracellular vestibule binding site. Two residues, Phe193 and Tyr308, are key energy barriers for the conformational changes. At the initial stage, the alprenolol is binding on the surface, where the Phe193 and Tyr308 close the tunnel (**Figure 5b**). As alprenolol moves deeper into the tunnel, Phe193 and Tyr308 undergo significant conformational changes to create an opening for entrance (**Figure 5c-d**). Once alprenolol reaches the orthosteric binding site, Phe193 and Tyr308 close the tunnel again to stabilize ligand binding (**Figure 5e**). In the second example, protein flexibility during 3T structure generation is particularly important because Phe193 and Tyr308 sidechain of the protein structure serve as the gateway for allowing or blocking access between the vestibular and the orthosteric binding sites.

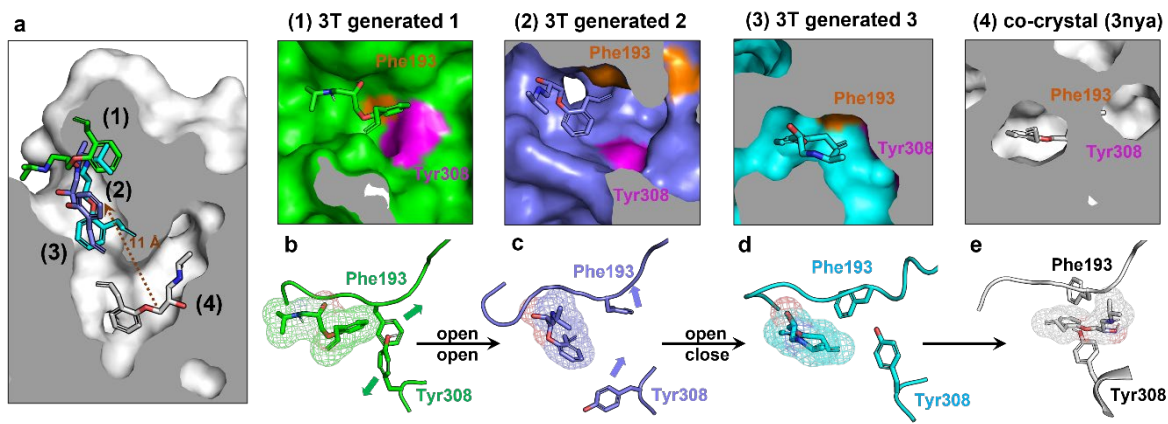

**Figure 5 | Exploration of protein-ligand conformations for distant poses and pocket locations**. **a)** the poses of alprenolol molecule in ADRB2 vestibular (1-3) and orthosteric receptor binding site (4) overlaid on the GPCR 3nya co-crystal. The vestibular binding site 3T poses (**b-d**) are generated using aggressive energetic kick from the original orthosteric site co-crystal pose (**e**). In the 3T conformations, we notice that these 3T protein-ligand conformations also correspond to the sequence of Phe193 and Tyr308 sidechain groups opening and closing the passageway to the orthosteric binding pocket. Ligand poses at the vestibular binding site were previously observed in long 5 µs MD simulation.[69]

## Conclusions

In summary, we demonstrate a novel algorithm tiered tensor transform (3T) to generate realistic complex multi-scale structures such as protein-ligand complex conformation. The structure generation works by using a combination of one example initial structure, a differentiable structure evaluation cost function, a hierarchical multi-scale tensor transformation sequence, and a random energetic kick for initial structural distortion. Using this method, we can generate unique protein-ligand complex conformations for a given protein target and a ligand molecule drug candidate. We demonstrate that these 3T conformations are useful for active ligand classification purposes. Features from ten 3T conformations significantly surpass features from hundreds of rigid protein conformations and can be generated with more than 80× lower computation cost *vs* comparable MD simulations on a GPU. We further demonstrate 3T's capability to explore distant protein pocket binding sites and intermediate protein-ligand pocket transition states. Due to 3T's modularity, adaptation onto other fields in physical sciences such as optical nanostructure or microfluidic structure generations/optimizations should be straightforward if a relatively low-cost structure evaluation cost function is available.

## Data Availability

Source data are provided with this paper. All the input protein and ligand input structures used in this work, as well as the raw output data for generating **Figure 3**, **Figure 4** and **Figure 5** are publicly available in Tencent Quantum Laboratory Github https://www.github.com/tencent-quantum-lab/3T. Only a small selection of the generated protein-ligand output structure examples is available in the same page due to the large data size (in total, over 200k protein-ligand structures are generated for this work). Some of the remaining output structures are available from the authors upon reasonable request.

## Code Availability

The 3T structure generation code, as well as pre-processing and post-processing code for RMSD analysis and active ligand classification tasks necessary for constructing **Figure 3**, **Figure 4** and **Figure 5** are publicly available in Tencent Quantum Laboratory Github https://www.github.com/tencent-quantum-lab/3T.

## Methods

**Ligand Structure Collection**

The 3D ligand structures for the "COCRY" dataset come from Protein Data Bank[1]. For these downloaded *pdb* files, all the water and solvent molecules were removed. Then, all these co-crystal structures are aligned to a reference protein structure (1fin in CDK2, 1uyg in HSP90 and 1ezq in FXA). The natural ligands were then extracted to get the "COCRY" dataset. The small molecules other than the "COCRY" dataset come from several sources (DUD-E, DEKOIS 2.0 and CSAR) as described in the work of Ricci-Lopez et al.[1–4] For the DUD-E and DEKOIS 2.0 datasets, the 3D structures have already been generated, which could be used for docking directly. For CSAR dataset, the 3D structures were generated using OpenBabel from the SMILES.[5]

**Input Structure Preparation**

A single protein (*pdb*) conformation structure is obtained from the Protein Data Bank (1fin for CDK2, 1uyg for HSP90 and 1ezq for FXA).[1] The ligand molecules are then docked onto the rigid protein using smina[6] with parameters of "—scoring=vinardo –factor=100 –num_modes=5 –exhaustiveness=16". The docked ligand is then extracted onto a standalone *mol2* file. Secondary structure, residue name and rotatable bond information are extracted from the *pdb* and *mol2* files using PyMol (v2.5), OpenBabel (v3.1.1) and RDKit (v2020.09.1.0).[5,7,8] The protein *pdb* is assigned CHARMM force field[9] using the GROMACS software[10] and converted into protein *gro* file. Similarly, CHARMM-style force field is assigned onto the ligand *mol2* file using SwissParam webserver,[11] which is then converted into the GROMACS format using charmm2gromacs-pvm functionality.[10] They are then combined onto one protein-ligand complex *gro* file (with complete CHARMM force field) and further converted into the LAMMPS input data format[12] using a custom version of the InterMol software[13] with some bug fixes. This LAMMPS input data format can be directly loaded onto our 3T PyTorch model for eventual atomistic force field energy (structure evaluation cost function) calculation. Secondary structure from the *pdb* files are assigned with PyMol. Here, we simply use three types of markers, i.e. helix, sheet and loop. Rotatable bond information is extracted from *mol2* files using OpenBabel and RDKit. $r_{centre}$ is calculated using the centre of mass of a batch of aligned cocrystal ligands.

**3T PyTorch Model Development and Structure Generation**

The 3T structure generation algorithm is implemented using the autograd functionality of PyTorch,[14] and computationally will look identical to a standard PyTorch deep learning model, except that there is no machine learning or training data involved in the process. The 3T model is split onto two parts, with the first being the hierarchical tensor transformation module where structural transformation happens and the second being the structure – force field energy calculation module. LAMMPS force field styles which are generated by InterMol are re-implemented in the 3T PyTorch model to enable native 'training-like' PyTorch structure generation. The energy cost function of the 3T system (for either $E_{complex,3T}$, $E_{ligand,3T}$, or $E_{protein,init}$) is simply the sum of Lennard-Jones, Coulombic, bond, angle, as well as proper and improper dihedral energies of the system snapshots during the 3T procedure for a given 3T model step the way they are usually calculated in LAMMPS.[12] Adam optimizer[15] and multi-step learning rate scheduler are used. In the beginning, 200 optimizer steps are used to relax the ligand structure in the pocket. Then the entire movable protein-ligand pocket (within $r_{cutoff}$) experiences 3T energetic kick, followed by 2000 3T optimizer steps. We use uniform random distribution $[-1.5, +1.5]$ Å for the micro-group $\theta_i$ translation kicks and $[-0.15, +0.15]$ radian for the $\theta_i$ rotation kicks. Some of the protein micro-groups such as phenylalanine, histidine, and tryptophan sidechains can be very rigid, which might introduce a very deep local structure energy minimum. Because of that, when we detect that there is enough space for these structures (no atomic clashes within 1 Å), we apply additional 180-degree rotation on $\theta_{A,i}$ with 50% probability during the 3T energetic kick step to enable more diverse protein-ligand complex pocket conformation generation. The generated conformations are outputted as *xyz* or *cif* files, and the cost function (energy landscape) throughout optimization was recorded. This process was repeated 10 times using 10 different (but consistent across ligands) random number seeds during the energetic kick, to generate ten 3T conformations. The PyToch components of 3T are executed on single NVIDIA T4 GPUs in the Tencent Cloud platform.

**PCA for Protein Structures**

Three groups of structures are processed for the analyses: (1) co-crystal protein structures, (2) protein conformations extracted from long MD simulations, and (3) 3T-generated protein conformations. First, all structures are aligned to the reference structures (e.g. 1fin for CDK2). The co-crystal protein conformations are taken from all available PDB's for the given protein, the MD conformations are 500 structures sampled every 1ns from a holo-MD simulation of the protein-ligand structures, while the 3T conformations are 90 structures generated from the 1fin smina re-docked initial structures. Then, the (x,y,z) coordinates of the protein backbone alpha carbon atoms in the pockets are extracted as features. Next, the scikit-learn PCA models (n_component=2) are fitted using the co-crystal data and then used to transform the MD and 3T data. Finally, the principle components PC1 and PC2 are plotted to show the protein conformation distributions.

**$\Delta RMSD$ Calculation**

The ten 3T conformations are scored using smina scoring function. Three pocket structures with the lowest docking scores are chosen as our best candidates, which are then aligned to the corresponding experimental co-crystals using PyMol based on the pocket atoms of the proteins, and the ligand RMSD is calculated with spyrmsd packages.[16] The best RMSD of the three aligned pocket structures is then compared to the initial smina pose' RMSD (similarly aligned to the experimental co-crystal). We also test how significant is the hypothesis that $\langle RMSD_{init} - RMSD_{3T} \rangle = \langle \Delta RMSD \rangle > 0$ on average, calculating the *p*-value using the scipy's stats package.

**Active Ligand Classification**

The recorded 3T formation energies (2200 steps for each conformation) are down sampled by 100 and scaled down by 1000 (to better match the unit of the docking scores), producing 22 energy features per conformation. The docking scores associated with initial structure, ligand-relaxed structure, and the 10 conformations are also included as features (12 features), resulting in a total of 232 features per ligand. These features are directly used as GBT-based 30×4cv active ligand classifier input, using the same Jupyter notebook available from previous work for AUC-ROC and $NEF_{R_a}$ calculations.[1] There is no change on the classification algorithm setup to ensure we have fair conformation feature comparison instead of classification algorithm comparison.

**Molecular Dynamics Setup and Semi-Flexible Docking Computation Resource Estimation**

Two types of machines are used for this comparison. For GPU machine, we use one NVIDIA T4 card. For CPU machine, we use 16 cores on a 48-core AMD EPYC 7K62 processor. Semi-flexible Smina docking (rigid protein backbone, rotatable sidechains) is done using parameters "--scoring=vinardo --factor=100 --num_modes=3 -exhaustiveness=16 --flexdist=4 --flexdist_ligand=ref_ligand.sdf". For rDock, the parameter "RECEPTOR_FLEX=4" is used in the PRMFILE and "-n=64" is used for docking. For the MD computation, the GROMACS protein-ligand structure above is solvated in water and the system is charge-neutralized and minimized before subsequent NVT and NPT equilibrations. GROMACS production MD speed (NPT, 2.0 fs time step at temperature of 300 K) is then measured. It is estimated that 1 μs to 1 ms MD time is needed to obtain enough protein structural diversity, and we have taken the aggressive assumption that 1 μs MD is enough. The required GPU computation cost is then calculated accordingly.[17]

# Method References

## Acknowledgements

The authors thank M. Shao from Tencent Quantum Lab for technical support on the Tencent Cloud platform. This work is fully conducted within Tencent Quantum Laboratory using the Tencent Cloud platform.

## Author Contributions

J.P.M. is responsible for 3T algorithm development, structure generation and feature extraction. Z.Y. and J.P.M. are responsible for initial protein-ligand complex structure preparation and input data pre-processing. Z.Y. is responsible for rigid protein docking, RMSD calculation and GBT classification. J.P.M and Z.Y. are responsible for MD structure generation and microstate analysis. Z.Y. and J.Q. are responsible for large-scale rigid protein docking on apo-MD structures. J.P.M. and Z.Y. performs the data and computation cost analysis. C.-Y.H. and S.Z. provide feedback and guide the research. All authors contribute into the manuscript preparation.

## Competing interests

The authors declare no competing interests.

## Supplementary Information

The online version contains supplementary material available at …

## Materials & Correspondence

Correspondence regarding this manuscript and material requests should be addressed to Jonathan Mailoa at [jpmailoa@alum.mit.edu](jpmailoa@alum.mit.edu) or Shengyu Zhang at [shengyzhang@tencent.com](shengyzhang@tencent.com).

# For Table of Contents Only

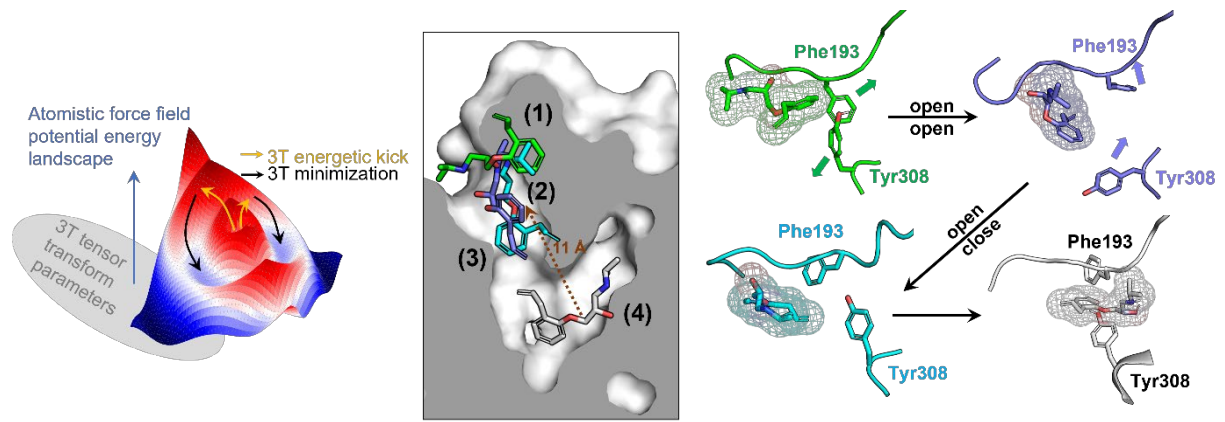

# Synopsis

We use multi-scale structure energy minimization to generate multiple fully-flexible protein-ligand pocket structures, enabling both accurate active ligand classification and distant ligand binding pose exploration.

# Supplementary Information –

# Protein-Ligand Complex Generator & Drug Screening via Tiered Tensor Transform

**Jonathan P. Mailoa,[1†*] Zhaofeng Ye,[1†] Jiezhong Qiu,[1] Chang-Yu Hsieh,[1] and Shengyu Zhang[2*]**

1) Tencent Quantum Laboratory, Tencent, Shenzhen, Guangdong, China
2) Tencent Quantum Laboratory, Tencent, Hong Kong SAR, China

**†** These authors contributed equally to this work

[*] corresponding author: jpmailoa@alum.mit.edu, shengyzhang@tencent.com

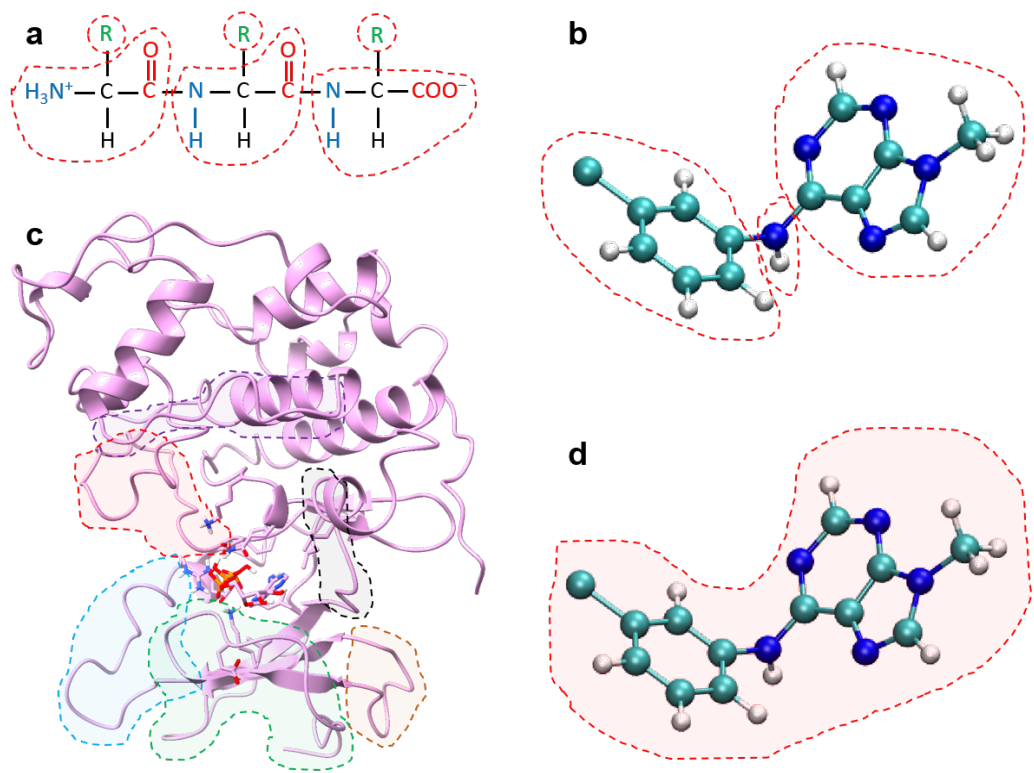

**Supplementary Figure 1 | 3T protein-ligand pocket hierarchical structure segmentation. a)** protein atom micro-groups based on amino acid backbone and sidechain, **b)** ligand atom micro-groups based on ligand rotatable bonds, **c)** protein atom macro-groups based on residue secondary structure, and **d)** ligand atom macro-group including all ligand atoms.

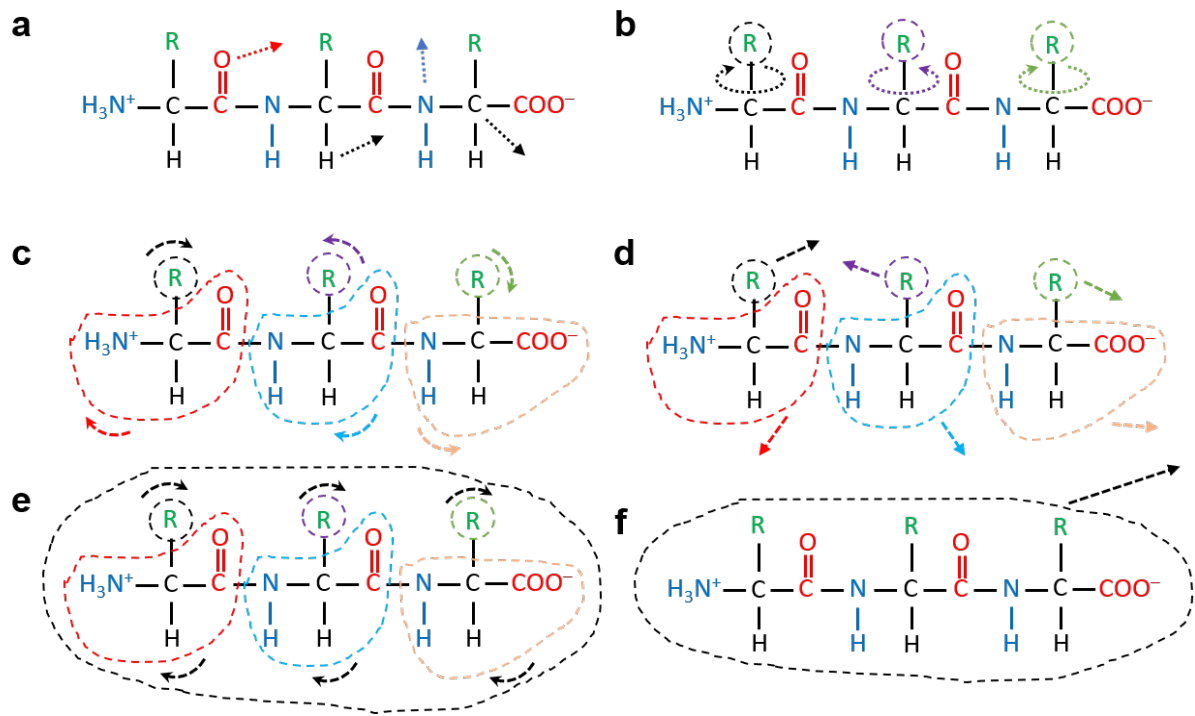

**Supplementary Figure 2 | Visual description of 3T hierarchical structure transformations. a)** Individual atom translation, **b)** individual micro-group sidechain rotations, **c)** individual micro-group centre rotations, **d)** individual micro-group translation, **e)** individual macro-group centre rotation, and **f)** individual macro-group translation. The parameters for these transformations are optimized by PyTorch force field-based cost function optimizers. For macro-group centre rotation, it is performed by rotating each micro-group within macro-group to its micro-group centre in a coordinated manner.

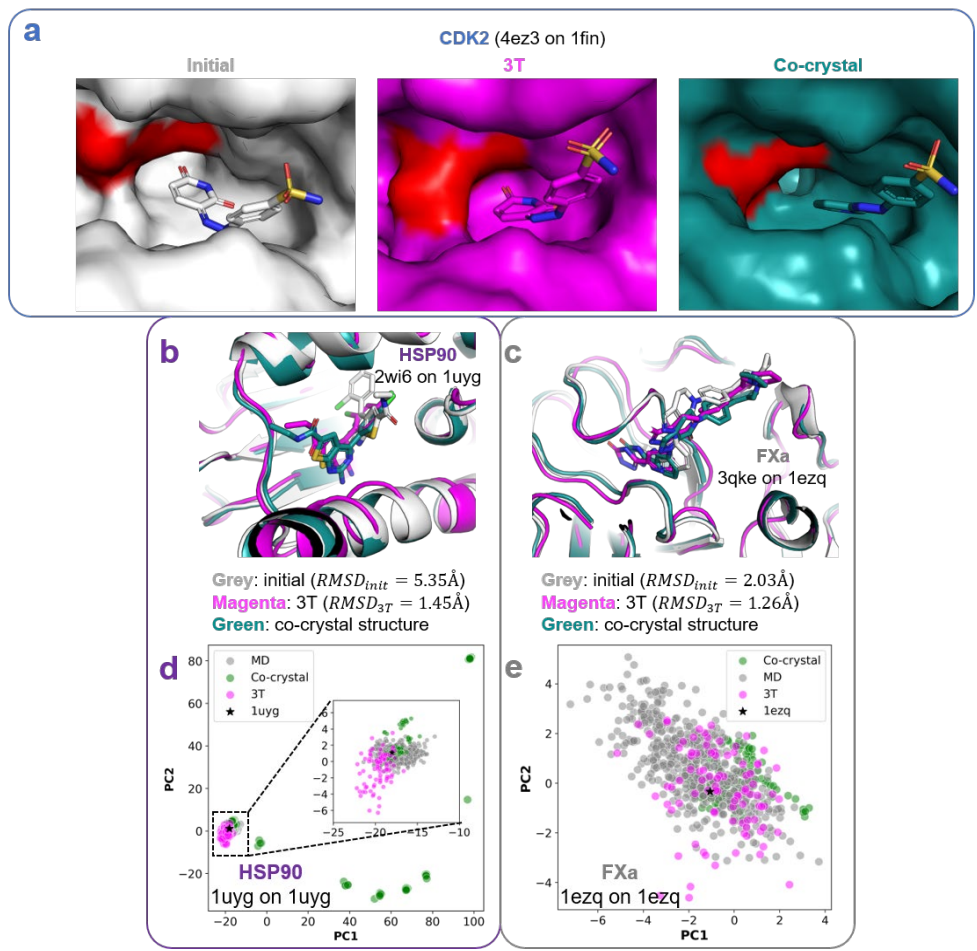

**Supplementary Figure 3 | Physical validation of 3T structures. a)** CDK2 ligand pose and protein surface mesh comparison for ligand from 4ez3 PDB cross-docked onto protein from 1fin, showing that the 3T transformation allows the protein pocket structure to contract and push the ligand to the right, similar to the actual 4ez3 co-crystal structure. The original 4ez3-1fin cross-docked structure has a rigid protein pocket which is more open, resulting in the wrong ligand pose. **b-c)** Individual example of overlaid structure comparison between cross-docked protein-ligand structure generated by 3T (magenta) vs initial cross-docked structure generated by smina (grey), as well as experimental co-crystal structure reference (green) for HSP90 (ligand of 2wi6 PDB cross-docked on protein of 1uyg PDB) and FXa proteins (ligand of 3qke PDB cross-docked on protein of 1ezq PDB). **d-e)** Protein backbone structure PCA comparison for all available experimental co-crystal PDB's (green), structures generated using a single holo-MD simulation (grey), and re-docked structures generated using 3T (magenta) for HSP90 and FXa proteins. The holo-MD and 3T re-docking are done using just HSP90 1uyg PDB and FXa 1ezq PDB, respectively. The PCA of the corresponding experimental co-crystal structures are highlighted in dark grey. For HSP90, there are several distinct protein-ligand pocket conformations available in nature (green). Holo-MD and 3T only occupy the PC subspace that belongs to the actual 1uyg PDB experimental co-crystal subspace and nowhere else. For FXa, it seems that there is only a single conformation space which is being shared by the co-crystals, holo-MD, and 3T. We note that we have much more limited availability of experimental FXa protein PDB co-crystals compared to CDK2 and HSP90 proteins.

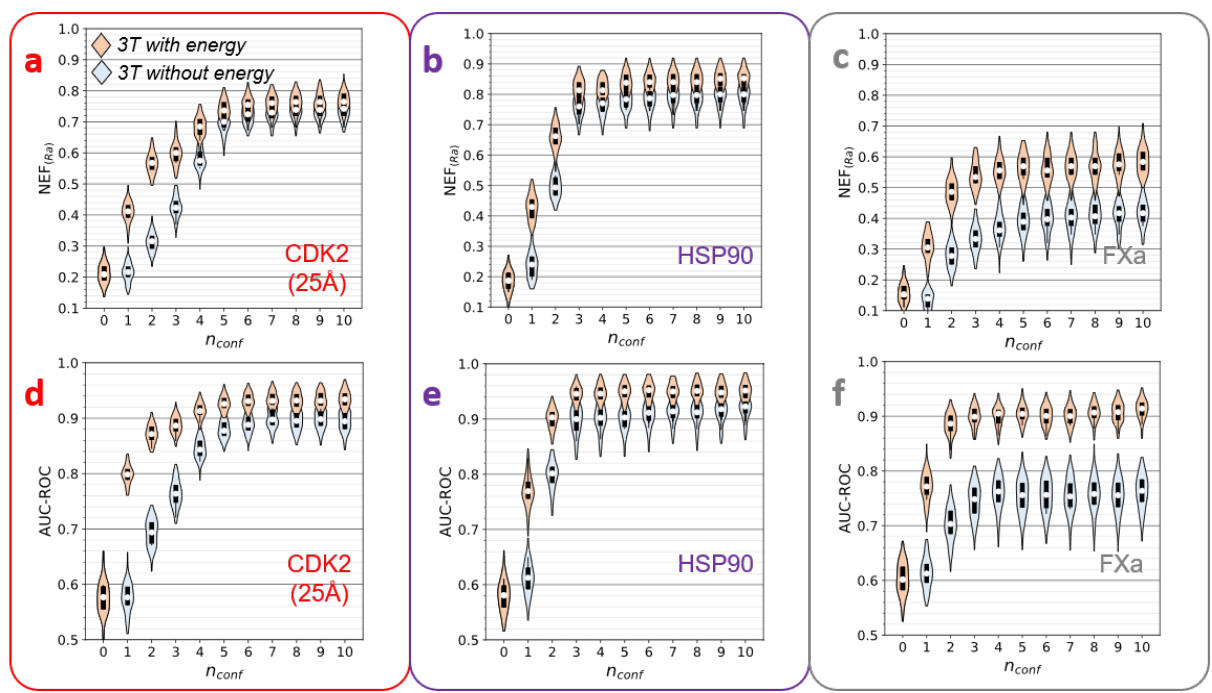

**Supplementary Figure 4 | The impact of incorporating 3T energy landscape features vs excluding such features during classification for different number of 3T-generated conformations. a–c)** The $NEF_{R_a}$ active ligand classification metric for CDK2 pockets ($r_{cutoff} = 25$Å), HSP90 pockets ($r_{cutoff} = 20$Å, non-rigid), and FXa pockets ($r_{cutoff} = 20$Å). **d–f)** AUC-ROC active ligand classification metric for the same proteins as in **a–c)**. 30× 4-fold cross validation (with GBT classifier) is used for the classification statistics. It can very clearly be seen that incorporating 3T energy landscape features is very helpful for identifying active ligands from the decoys. $n_{conf} = 0$ means that we do not generate any fully-flexible protein-ligand pocket structure through 3T energetic kick, and only utilize the initial smina cross-docked structure and the structure obtained after relaxing the ligand while maintaining rigid protein pocket (see **Figure 4**).

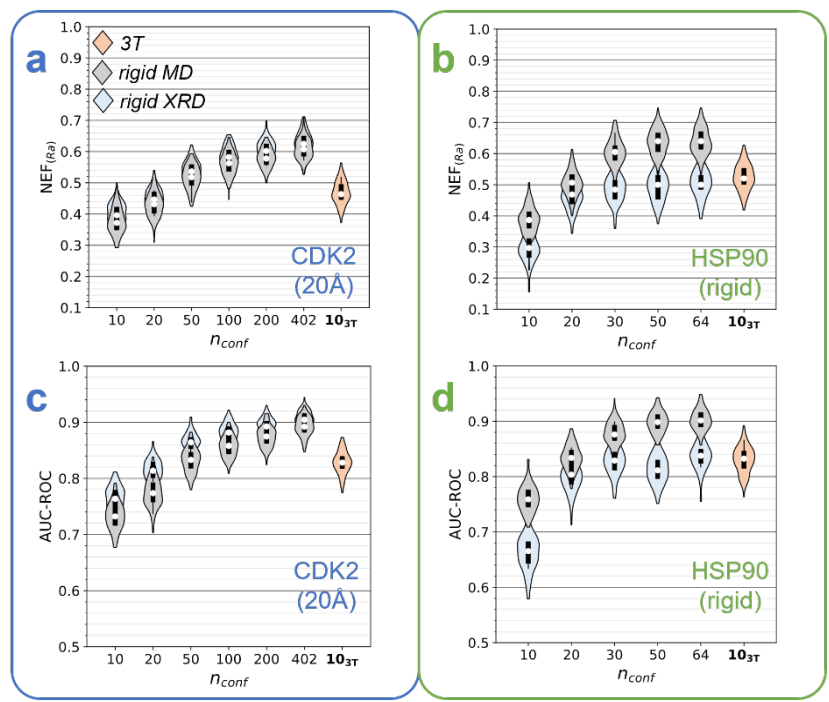

**Supplementary Figure 5 | Impact of reducing or eliminating 3T protein flexibility on active ligand classification performance. a–b)** The $NEF_{R_a}$ active ligand classification metric for CDK2 pockets ($r_{cutoff} = 20\text{Å}$) and HSP90 pockets (pockets intentionally made to be perfectly rigid and non-transformable by 3T), and **c-d)** AUC-ROC active ligand classification metric for the same proteins. 30× 4-fold cross validation (with GBT classifier) is used for the classification statistics. It can be seen that 3T on small CDK2 pocket is insufficient to build good classifiers and a larger radius of pocket flexibility needs to be allowed. Similarly, it can also be seen that performing 3T just on the ligands (with perfectly rigid HSP90 protein pocket structure) will significantly degrade the active ligand classifier performance.

| Protein | Active Ligands | | 3T Classifier | Rigid MD Classifier | Rigid XRD Classifier | $n_{conf,3T}$ vs $n_{conf,MD-XRD}$ |
|---|---|---|---|---|---|---|
| CDK2 (25 Å) | 442/3764 ($R_a = 0.117$) | $NEF_{R_a}$ | **0.771 ± 0.030** | 0.608 ± 0.033 | 0.624 ± 0.039 | **(10 + 2)** vs 402 |
| | | AUC-ROC | **0.935 ± 0.014** | 0.892 ± 0.017 | 0.904 ± 0.015 | |
| CDK2 (20 Å) | 442/3764 ($R_a = 0.117$) | $NEF_{R_a}$ | 0.469 ± 0.037 | 0.608 ± 0.033 | **0.624 ± 0.039** | (10 + 2) vs **402** |
| | | AUC-ROC | 0.828 ± 0.020 | 0.892 ± 0.017 | **0.904 ± 0.015** | |
| HSP90 | 298/2452 ($R_a = 0.122$) | $NEF_{R_a}$ | **0.851 ± 0.035** | 0.640 ± 0.042 | 0.505 ± 0.046 | **(10 + 2)** vs 64 |
| | | AUC-ROC | **0.949 ± 0.018** | 0.903 ± 0.019 | 0.836 ± 0.024 | |
| HSP90 (rigid) | 298/2452 ($R_a = 0.122$) | $NEF_{R_a}$ | 0.524 ± 0.042 | **0.640 ± 0.042** | 0.505 ± 0.046 | (10 + 2) vs **64** |
| | | AUC-ROC | 0.830 ± 0.024 | **0.903 ± 0.019** | 0.836 ± 0.024 | |
| FXa | 298/7191 ($R_a = 0.040$) | $NEF_{R_a}$ | **0.584 ± 0.043** | 0.554 ± 0.046 | 0.452 ± 0.044 | (10 + 2) vs 136 |
| | | AUC-ROC | **0.913 ± 0.018** | 0.902 ± 0.021 | 0.855 ± 0.021 | |

**Supplementary Table 1 | 3T active ligand classification metrics using the GBT classifier across 3 different protein hosts.** In addition to the information available in **Table 1** of the main text, we have also included the classification statistics when $r_{cutoff} = 20$Å pocket is used for CDK2 and when perfectly rigid HSP90 protein pocket is used for 3T protein-ligand pocket conformation generations. It can be seen that when the flexible protein pocket generation ability of 3T is reduced or removed, its active ligand classification utility will be diminished.

| Protein | Protein Rigidity | Method | XRD Experiment | MD Cost / Ligand | Docking Cost / Ligand |
|---|---|---|---|---|---|
| CDK2 | Rigid | Rigid XRD | 406× | 0 | 12.9 CPU-hr |
| | | Rigid MD* | 1× | 0.058 GPU-hr | 12.9 CPU-hr |
| | Semi-flexible | Smina (flexible sidechain) | 406× | 0 | 710.1 CPU-hr |
| | | rDock (flexible OH, NH$_3$) | 406× | 0 | 6.3 CPU-hr |
| | Fully-flexible | Holo-MD* | 1× | 217.2 GPU-hr | 0 |
| | | 3T | 1× | 0 | 2.5 GPU-hr / 22.7 CPU-hr |
| | * 69k atoms for the CDK2 protein --> GROMACS GPU speed = 110.5 ns/day ||||||
| HSP90 | Rigid | Rigid XRD | 64× | 0 | 4.1 CPU-hr |
| | | Rigid MD* | 1× | 0.051 GPU-hr | 4.1 CPU-hr |
| | Semi-flexible | Smina (flexible sidechain) | 64× | 0 | 178.3 CPU-hr |
| | | rDock (flexible OH, NH$_3$) | 64× | 0 | 1.7 CPU-hr |
| | Fully-flexible | Holo-MD* | 1× | 124.3 GPU-hr | 0 |
| | | 3T | 1× | 0 | 2.1 GPU-hr / 10.1 CPU-hr |
| | * 37k atoms for the HSP90 protein --> GROMACS GPU speed = 193.1 ns/day ||||||
| FXa | Rigid | Rigid XRD | 136× | 0 | 6.5 CPU-hr |
| | | Rigid MD* | 1× | 0.028 GPU-hr | 6.5 CPU-hr |
| | Semi-flexible | Smina (flexible sidechain) | 136× | 0 | 170.5 CPU-hr |
| | | rDock (flexible OH, NH$_3$) | 136× | 0 | 2.5 CPU-hr |
| | Fully-flexible | Holo-MD* | 1× | 201.7 GPU-hr | 0 |
| | | 3T | 1× | 0 | 1.8 GPU-hr / 10.7 CPU-hr |
| | * 60k atoms for the FXa protein --> GROMACS GPU speed = 119.0 ns/day ||||||

**Supplementary Table 2 | Experimental and computational resource estimation for various docking-based active ligand classification tasks on the CDK2, HSP90, and FXa dataset based on the protein-ligand complex structure generation method being used.** Active ligand Experimental and computational resource estimation for various docking-based active ligand classification tasks on the CDK2, HSP90, and FXa dataset, based on the protein-ligand complex structure generation method being used. The six methods are categorized based on the rigidity of generated protein structure (rigid: no protein conformation change during ligand cross-docking, semi-flexible: some protein sidechain rotation is allowed during cross-docking, fully-flexible: all protein backbone and sidechain atoms can freely move during cross-docking). The three methods in bold are the ones for which we do classification performance comparison in main text **Figure 4**. The XRD experiment column refers to the number of X-ray diffraction co-crystal structures which are needed to enable active ligand classification task based on such methods (the same number of experimental structures being used in main text **Figure 4**). For rigid MD, holo-MD, and 3T, only one such experimental co-crystal protein structure is needed as the initial structure. However, the holo-MD and rigid MD methods will require additional MD simulation cost for the subsequent protein-ligand complex structure generations. For 3T, all structure generation cost (single rigid protein docking plus 10-conformation generation) is categorized as 'docking cost'. The docking computation cost estimates are averaged from three randomly chosen ligands, except for 3T CPU-hr estimates which are averaged from 16 randomly chosen ligands.

*Rigid XRD, Smina (flexible sidechain) & rDock (flexible OH, NH₃)*

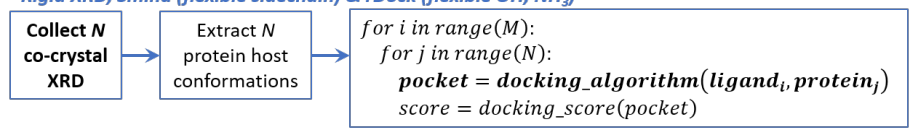

*Rigid MD*

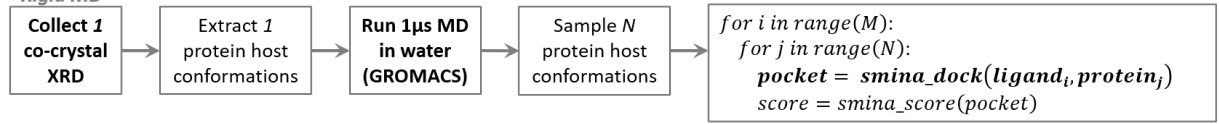

*Holo-MD*

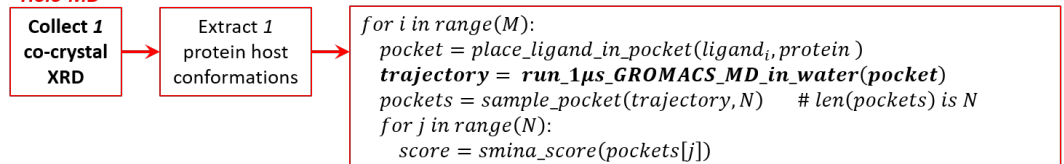

*3T*

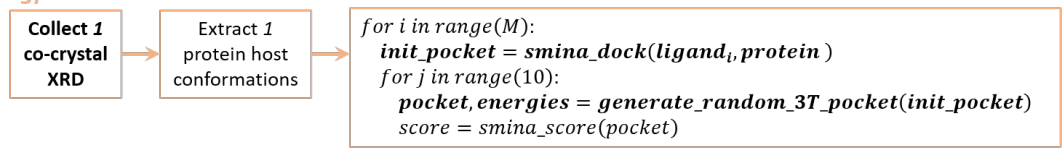

**Supplementary Flowchart 1 | Pseudocode and flowchart for various docking-based active ligand classification tasks described in Supplementary Table 2.** For CDK2, HSP90, and FXa proteins, the value of *N* is 406, 64, and 136 respectively. The docking algorithm used for rigid XRD is a standard smina docking (protein host structure frozen) algorithm. Sub-tasks which are experimentally or computationally demanding are bolded. The timing/ligand metric in **Supplementary Table 2** is average resource requirement for each $ligand_i$. The docking score calculation (with no conformation search) and classification model training time itself is negligible compared to these feature collection tasks.

## Significantly More Aggressive 3T Energetic Kick

In the main text, we use relatively gentle energetic kick settings to screen the 10-20k protein-ligand complexes while ensuring that 3T does not end up generating significant numbers of non-physical protein-ligand complex conformations. In this section, we offer an alternative scheme where we can take more risk and utilize significantly more aggressive energetic kick settings when working on much smaller number of protein-ligand samples of interest. We apply this new setting on one difficult problem, which is the exploration of distant metastable binding poses within a protein pocket such as an alprenolol ligand docked into the beta2 GPCR receptor, with a different vestibular binding site located relatively far (approximately 10-15 Å away from the receptor pocket). We choose three initial structures, which are the 1st to 3rd-ranked smina-docked poses. We are interested in seeing 3T explore different conformations within the vestibular binding site (no experimental co-crystal available for experimental comparison because this is just a temporary transition site, as identified by lengthy 5 μs molecular dynamics work by Dror *et al*, doi:10.1073/pnas.1104614108), starting from its initial orthosteric binding site location.

To enable these tasks, we perform the following aggressive modifications:

a. Right before the first ligand relaxation step (main text **Figure 4a**), we perform 3T energetic kick for the ligand's macro-group transform parameters. We set $\theta_{R,j}$ uniformly between $[-\pi, \pi]$ for rotation around the $x$, $y$, and $z$ axis while initial $\theta_{T,j}$ value is set to $[-3, -9, -9]$ which is around the general approximate direction of the vestibular binding site. This must be done because we currently have no automated pocket identification functionalities, and we want to prevent 3T from kicking the ligand into random dense protein volumes (it is inefficient to make 3T recover the structure if the initial structure is completely unphysical, such as ligands entangled by multiple protein components). It is computationally unwise to let 3T perform random and large macro-group translation energetic kick in this aggressive case.

b.  The ligand macro-group no longer follows the rotation mode of protein macro-groups (**Supplementary Figure 2e**) which is a coordinated rotation of micro-groups with respect to individual micro-group centres. The ligand macro-group will rotate with respect to the centre of the entire ligand macro-group instead. This enables a more mobile ligand, while keeping the protein structure transformation more conservative.

c.  The PyTorch optimizer settings for the ligand relaxation step is increased by 100×.

d.  The gradient of 3T parameters ($\vec{r}_{m,init}$ and $\theta$'s) for the ligand is multiplied by 3×, while the protein 3T parameters are not modified. This ensures that ligands can be adjusted even faster, while the protein structure transformation is kept more conservative.

e.  The initial ligand relaxation step is increased to 10,000 steps (vs the initial 200 steps). The full pocket relaxation step following the second energetic kick is kept at 2,000 steps.

f.  We generate 30 conformations per protein-ligand pocket complex pair instead of the main text's setting of 10 conformations per pair. This means that we initially generate 90 conformations.

g.  After 2,000 steps of pocket relaxation, we choose 10 structures from the 90 conformations and extend the full pocket relaxation step from 2,000 steps to 5,000 steps.

h.  1 structure out of the 10 structures was not fully relaxed after 5,000 steps and need to undergo an additional 5,000 full pocket relaxation steps to further relax the protein structures into their final conformations.

Overall, it took approximately 55 GPU-hr of computation time in one Nvidia T4 GPU machine for all the conformations we explored.